\documentclass{article}
\usepackage{enumitem}   
\usepackage{spconf,amsmath,graphicx,multirow}
\usepackage{multirow}
\usepackage{tablefootnote}

\title{KL-Divergence-Based Region Proposal Network for Object Detection}
\name{Geonseok Seo$^{1 2}$\sthanks{The work was done at Seoul National University.} \qquad Jaeyoung Yoo$^{1}$ \qquad Jaeseok Choi$^{1}$ \qquad Nojun Kwak$^{1}$\sthanks{Corresponding author. This work was supported by the ICT R\&D program of MSIP/IITP, Korean Government(2017-0-00306).}}
  
\address{$^{1}$ Department of Intelligence and Information, Seoul National University, South Korea \\
      $^{2}$ Samsung Advanced Institute of Technology (SAIT), South Korea}

\begin{document}
%
\maketitle

\begin{abstract}
The learning of the region proposal in object detection using the deep neural networks (DNN) is divided into two tasks: binary classification and bounding box regression task. However, traditional RPN (Region Proposal Network) defines these two tasks as different problems, and they are trained independently. In this paper, we propose a new region proposal learning method that considers the bounding box offset's uncertainty in the objectness score. Our method redefines RPN to a problem of minimizing the KL-divergence, difference between the two probability distributions. We applied KL-RPN, which performs region proposal using KL-Divergence, to the existing two-stage object detection framework and showed that it can improve the performance of the existing method. Experiments show that it achieves 2.6\% and 2.0\% AP improvements on MS COCO test-dev in Faster R-CNN with VGG-16 and R-FCN with ResNet-101 backbone, respectively. 
\end{abstract}
\vspace{-2mm}
\section{Introduction}
\label{sec:intro}
In the field of computer vision, object detection has been an important problem for a long time. Now, object detection using DNN has shown excellent performance and has been used in many industrial fields like face recognition \cite{zhang2017s3fd,wang2017detecting}, pedestrian detection \cite{dollar2009pedestrian,zhang2018occluded}, and autonomous vehicle \cite{lin2017fast,hu2018sinet}. 
Object detection is mainly divided into the single-stage method and the two-stage method. The first step of the two-stage method is the region proposal. 
To find region proposal, Faster R-CNN \cite{ren2015faster} uses RPN and it reduces computation and improves performance compared to methods that do not use neural networks. Currently, RPN based two-stage object detection methods \cite{lu2019grid,he2017mask,cai2018cascade,dai2016r,lin2017feature} show high performance.
\begin{figure}[t]
\begin{center}
\includegraphics[width=1.0\linewidth]{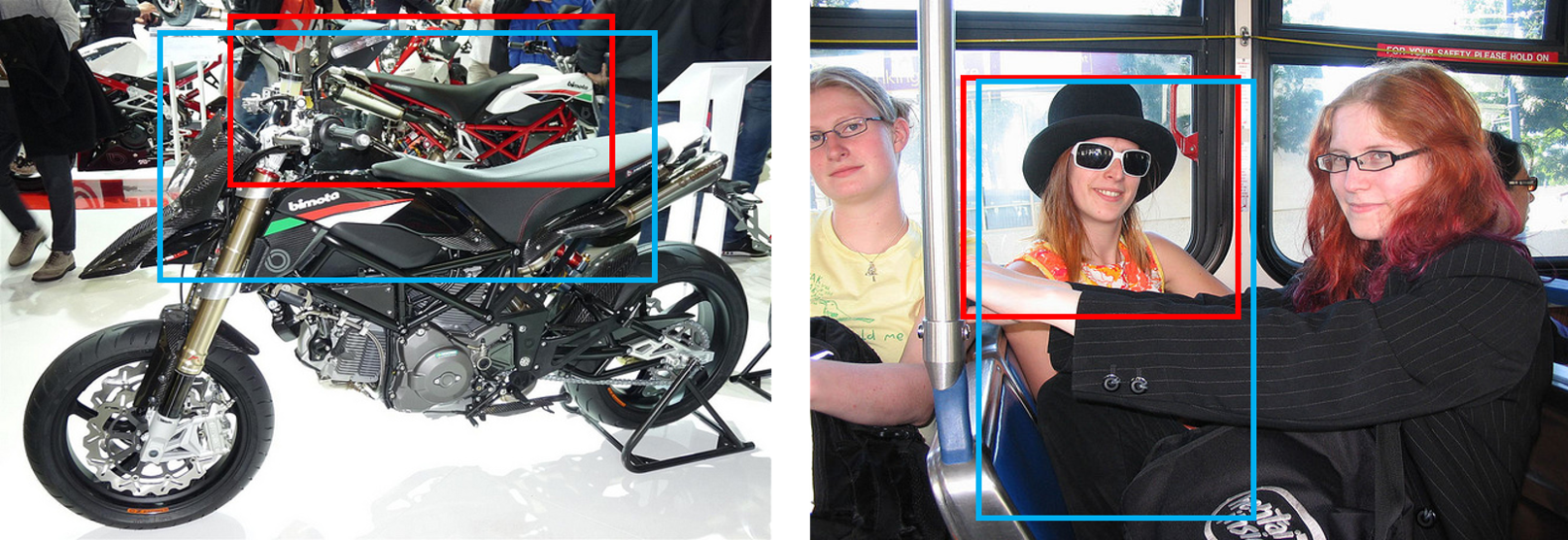}
\end{center}\vspace{-4mm}
   \caption{Some cases of objectness score relation with uncertainty of coordinates. In both cases, the objectness score of the blue boxes should be lower than the red boxes considering high uncertainty of coordinate because of occlusion.}\vspace{-3mm}
\label{fig_ob_coord_uncertainty}
\end{figure}
For training RPN, two loss functions are used. First is a binary cross-entropy loss that classifies candidate regions as foreground or background. Second is the smooth $L_{1}$ loss that used to learn the coordinate offset. 
It applies to $L_{2}$ loss for small offset and $L_{1}$ loss for large offset.

\begin{figure*}[t]
\begin{center}
\includegraphics[width=1.0\linewidth]{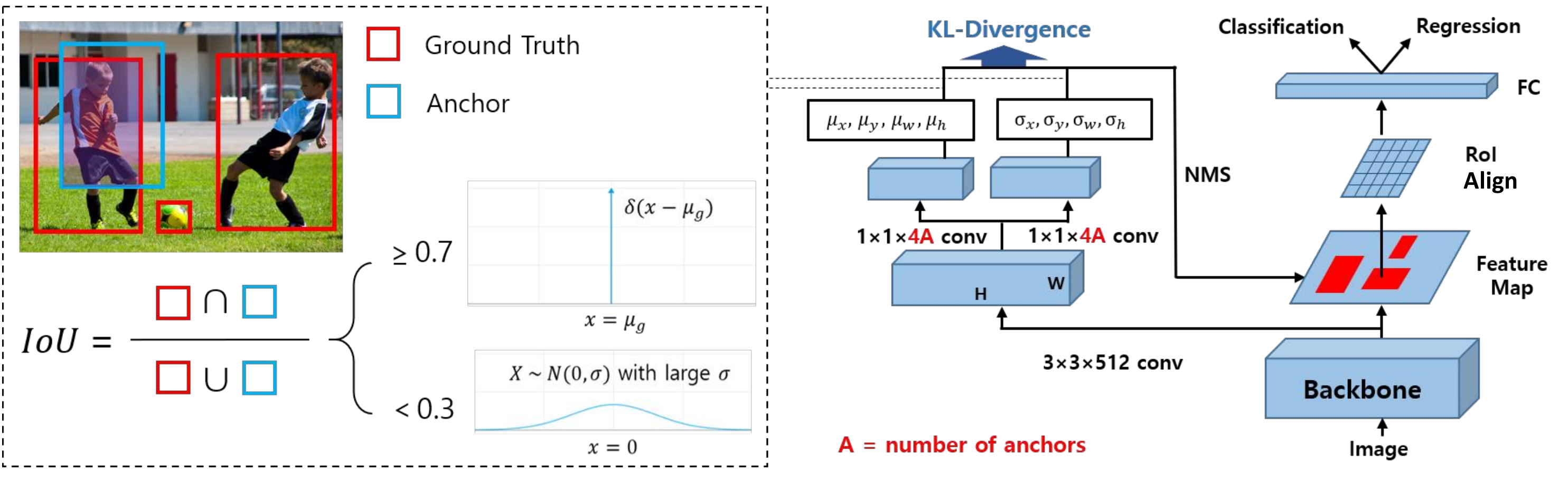}
\end{center}\vspace{-3mm}
   \caption{Network architecture of KL-RPN based Faster R-CNN. KL-RPN predicts the mean and standard deviation of bounding box offset. The target of example is determined based on IoU with ground truth bounding box.}\vspace{-3mm}
\label{fig:network}
\end{figure*}

However, there are some things that RPN does not consider. In general, because of occlusion and boundary ambiguities, ground truth bounding boxes have variations in the process of human annotation. In other words, the ground truth bounding box will be expressed as a probability distribution that some probability exists in the corresponding coordinate \cite{he2019bounding}. Since RPN uses the Smooth $L_{1}$ loss to learn only bounding box offset, it does not consider the uncertainty, the confidence that the coordinate exists, of the bounding box coordinate. The Smooth $L_{1}$ loss is known to be less sensitive to outliers than the normal $L_{2}$ loss, but ultimately it is not a regressor that takes uncertainty into account. In order to learn the uncertainty of coordinates, \cite{he2019bounding} used KL-Divergence loss and reflected it in the refinement of coordinates in NMS post-processing. In addition, in the existing RPN, two tasks, classification and bounding box regression, are learned independently and they do not concern each other in the training process. Because predicting region proposal is a matter of estimating the bounding box coordinates where the object is likely to exist, they are not independent and the classification should be considered together with bounding box regression for accurate modeling. In terms of the uncertainty of coordinates, Figure~\ref{fig_ob_coord_uncertainty} shows the several cases of objectness score relation with the uncertainty of coordinates. In the picture, the blue box has a high uncertainty of coordinates than the red box because of occlusion. In this case, the blue box's objectness score should be low. Therefore, the objectness score not only distinguishes whether the object is foreground or background but also should be taken into account the uncertainty of the coordinates. From this fact, unlike the existing RPN, we propose a new region proposal learning method named KL-RPN that considers the bounding box offset's uncertainty in the objectness score by using only a single loss function.

When modeling bounding box offset as a Gaussian distribution, the KL-RPN uses the KL-Divergence, the distance between these two probability distributions, for region proposal learning. 
Then we gave the probability distribution of each example as the target distribution that the network should learn. A positive example is a Dirac delta function with zero standard deviation and mean as an offset between ground truth bounding box and anchor. A negative example is assumed to be a Guassian distribution with large standard deviation so as to deem uniform distribution which has same probability at all values. 
Then we use this standard deviation as the objectness score, which applied to the post-processing NMS stage, to apply the bounding box offset’s uncertainty.

We applied our method to Faster R-CNN and R-FCN which are the two most commonly used two-stage object detection methods based on VGG-16 and ResNet-101 backbone and confirmed the performance improvement in PASCAL VOC and MS COCO datasets.

Our main contributions are summarized as follows:
\vspace{-2mm}
\begin{enumerate}[label=(\roman*)]
\item In RPN, objectness score and bounding box offset are trained independently. In KL-RPN, both of them are trained together by minimize KL-Divergence.
\vspace{-2mm}
\item In KL-RPN, bounding box offset’s uncertainty is considered as the objectness score.
\vspace{-2mm}
\item Our method adds little parameters to the existing RPN, so the computation is almost unchanged and also performance is improved, without bells and whistles.
\end{enumerate}


\vspace{-3mm}
\section{KL-RPN}
\label{sec:format}

\subsection{Target Distribution}
\label{sec:3_1}
Figure~\ref{fig:network} is the overall KL-RPN network structure. The network learns the $\mu$, and $\sigma$ of single variate Gaussian on the target distribution of $x$, $y$, $w$, and $h$ coordinate offsets. Like the existing RPN \cite{ren2015faster}, the target of example to be learned for each feature map is determined based on IoU (Intersection over Union) with ground truth bounding box. The anchors that IoU with any ground truth bounding box is 0.7 or more and the anchors with the highest IoU with a ground truth bounding box are positive, and negative for IoU with ground truth bounding box less than 0.3. Once the types of examples are determined, as shown in Figure~\ref{fig:network}, the KL-RPN uses a Dirac delta function with a $\sigma$ of zero as a target for positive example and a Gaussian distribution with large standard deviation as a target for negative example. In the negative example, it would be difficult for the network to learn a uniform distribution with an infinite standard deviation. Instead of uniform distribution, we set a Gaussian distribution with larger standard deviation  than that of positive example as a target for a negative sample to reflect high uncertainty. 
The target distribution mean $\mu$ of the positive example is the offset of $x$, $y$, $w$, and $h$ as in \cite{girshick2014rich,ren2015faster}.
\vspace{-3mm}
\subsection{KL-Divergence between Two Gaussian Distribution}
\label{sec:3_2}

In this section, we will introduce $L_{KL}$, used in KL-RPN learning. $L_{KL}$ is a loss function based on Kullback-Leibler Divergence. The Kullback-Leibler Divergence, KLD, is a function used to calculate the difference between two probability distributions. The difference between the target probability distribution $P(x)$ and the probability distribution $Q(x)$ predicted by the network can be obtained by using the KLD. The expression of the KLD is defined as follows.
\begin{equation}
\begin{aligned}
\label{eq:kld}
    D_{KL}(P||Q) &=  E_{x\sim P}\left[\log\frac{P(x)}{Q(x)}\right], \\ &= E_{x\sim P}\left[\log P(x)-\log Q(x)\right]
\end{aligned}
\end{equation}
Since we want to minimize the difference between the $P(x)$ and the $Q(x)$, the loss function of the KLD is:
\vspace{-0.5mm}
\begin{equation}\begin{aligned}\label{eq:kld2}
    L &= D_{KL}(P(x)||Q(x)), \\ &= \int P(x)\log P(x)\, dx - \int P(x)\log Q(x)\, dx
\end{aligned}
\end{equation}

We assume that the $P(x)$ and the $Q(x)$ have a Gaussian distribution. In this case, the KLD of the two Gaussian distributions is calculated as follows. Let $P(x) = \mathcal{N}(\mu_{1}, \sigma_{1})$ and $Q(x) = \mathcal{N}(\mu_{2}, \sigma_{2})$. From the equation (\ref{eq:kld2}), the loss function of the overall KLD is:
\begin{equation}
\begin{aligned}
\label{eq:total_kd}
L&=\int P(x)\log P(x)\, dx - \int P(x)\log Q(x)\, dx \\ 
&=\frac{\sigma_{1}^2+(\mu_{1}-\mu_{2})^2}{2\sigma_{2}^2}+\log\frac{\sigma_{2}}{\sigma_{1}}-\frac{1}{2}
\end{aligned}
\end{equation}

When $P(x)$ is a negative sample, we regard it as Gaussian distribution with $\mathcal{N}(\mu_{1}=0, \sigma_{1}^2=\frac{1}{3})$. And We predicted the $\beta = log(\sigma_{2}^2)$ as \cite{he2019bounding} to prevent gradient exploitation of $\sigma$. Considering only the terms related to the gradient, we can rearrange the loss equation as follows.
\vspace{-1.0mm}
\begin{equation}
\begin{aligned}
\label{eq:Lbg}
    L_{neg}=\frac{e^-\beta (\gamma \cdot \mu_{2}^2+\frac{1}{3})}{2}+\frac{\beta}{2}
\end{aligned}
\end{equation}
In here, we set the $\gamma = \frac{1}{2}$.
\vspace{-1.0mm}
\subsection{KL-Divergence between Dirac Delta Function and Gaussian Distribution}
\label{method:kd_div_b_gd_and_ddf}
As mentioned above, we regard the distribution of positive examples as a Dirac delta function with a standard deviation of zero as in \cite{he2019bounding}.
\vspace{-1mm}
\begin{equation}
\begin{aligned}
\label{eq:dirac delta}
    P(x) = \delta(x-\mu_{1})
\end{aligned}
\end{equation}
Here, $\mu_{1}$ is the offset between the ground truth bounding box and anchor. Now we will discuss the KLD between the Dirac delta function and the Gaussian distribution. In equation (\ref{eq:kld2}), the probability exists only at $x = \mu_{1}$, the KLD is:
\vspace{-0.5mm}
\begin{equation}
\begin{aligned}
\label{eq:Dirac-Gaussian}
    L&=\frac{(\mu_{1}-\mu_{2})^2}{2\sigma_{2}^2}+\frac{1}{2}\log(2\pi\sigma_{2}^2)-H(P(x))
\end{aligned}
\end{equation}
In here, the $H(P(x))$ is entropy of $P(x)$. By leaving the term related to the gradient, the loss equation can be summarized as follows. 
\vspace{-1mm}
\begin{equation}
\begin{aligned}
\label{eq:Lfg}
    L_{pos}=\frac{e^-\beta \gamma \cdot(\mu_{1}-\mu_{2})^2}{2}+\frac{\beta}{2}
\end{aligned}
\end{equation}
In here, we set the $\gamma = \frac{1}{2}$.
\vspace{-2mm}
\subsection{Entire Loss Function}
\label{method:entire_loss}


As described in Section \ref{sec:3_1}, for each example, the target to be learned is determined based on IoU between ground truth bounding box and anchor. In each case, the loss function is described as follows.\vspace{-0.5mm}
\begin{equation}
\begin{aligned}
\label{eq:Lkl}
    L_{KL}=\begin{cases}
L_{pos}, & \mbox{For positive example} \\
L_{neg}, & \mbox{For negative example}
\end{cases}
\end{aligned}
\end{equation}
After the RoIs through KL-RPN are detected, we pool the RoIs from KL-RPN and finally perform classification and regression on each RoIs as in \cite{ren2015faster,dai2016r}. In this stage, the loss functions are $L_{cls}$ and $L_{reg}$, respectively. The total loss equation for training the entire object detection network is:
\begin{equation}
\begin{aligned}
\label{eq:Ltotal}
    L = L_{KL}+\beta\cdot (L_{cls}+L_{reg})
\end{aligned}
\end{equation}
In here, we set the $\beta$ = 10.
\vspace{-2.5mm}
\subsection{Objectness Score by using Standard Deviation}
\label{sec:objectness_score}
We use the network's standard deviation as the objectness score to apply the uncertainty of the bounding box offset to the objectness score. Like equation (\ref{eq:cls_score}), the objectness score is defined as reciprocal of multiplication of the standard deviation of a single variate Gaussian whose variables are $x$, $y$, $w$, and $h$.
\vspace{-1mm}
\begin{equation}
\begin{aligned}
\label{eq:cls_score}
Class_{score} = \frac{1}{\sigma_{2 \_x} \cdot \ \sigma_{2 \_y} \cdot \ \sigma_{2 \_w} \cdot \ \sigma_{2 \_h} + \epsilon}
\end{aligned}
\end{equation}
Here, $\sigma_{2}$ is the standard deviation that the network predicted and $\epsilon$ was set to $1e^{-12}$ to prevent division by zero. Following the section \ref{sec:3_2} and \ref{method:kd_div_b_gd_and_ddf}, the $\sigma_{2}$ for the $x$, $y$, $w$, and $h$ offsets will be close to zero for a positive example and close to $\sigma_{1}$ for a negative example. Therefore, the lower $\sigma_{2}$, the better positive sample can be considered. When the RoIs predicted by KL-RPN come out, the classification and regression of the second stage of the \cite{ren2015faster,dai2016r} are performed by doing the NMS based on the objectness score in the order of bigger. In this part, training is performed in the same manner as in \cite{ren2015faster,dai2016r}.

\vspace{-2mm}
\section{Experiments}
\label{sec:pagestyle}

We applied KL-RPN to Faster R-CNN and R-FCN, both of which are two-stage object detection method used RPN. \cite{yang2017faster} was used as the baseline of the experiment, and used VGG-16 \cite{simonyan2014very} and ResNet-101 \cite{he2016deep} backbone pretrained with ImageNet \cite{deng2009imagenet}. PASCAL VOC \cite{everingham2010pascal} and MS COCO \cite{lin2014microsoft} dataset were used for training. In all experiments, we used a PyTorch \cite{paszke2017automatic}.

\begin{table*}[t]
\begin{center}
\scalebox{0.86}{
\begin{tabular}{|c|c|c|c|c|c|c|c|c|c|c|c|c|c|}
\hline
Method                        & train data                   & bs & lr   & \begin{tabular}[c]{@{}c@{}}Region\\ Proposal\end{tabular} & Backbone                    & AP            & AP$_{50}$          & AP$_{75}$          & AP$_{S}$           & AP$_{M}$           & AP$_{L}$           & Param(M) & Speed(ms) \\ \hline
\multirow{4}{*}{Faster R-CNN} & \multirow{4}{*}{trainval35k} & 16  & 1e-2 & RPN\cite{yang2017faster}                                                       & \multirow{2}{*}{VGG-16}     & 27.1          & 47.1          & 27.7          & 11.5          & 30.0          & 36.5          & 138.31   & 188       \\ \cline{3-5} \cline{7-14} 
                              &                              & 16  & 6e-5 & KL-RPN                                                    &                             & \textbf{29.7} & \textbf{49.9} & \textbf{31.5} & \textbf{11.6} & \textbf{33.0} & \textbf{42.4} & 138.32   & 187       \\ \cline{3-14} 
                              &                              & 16  & 1e-2 & RPN\cite{yang2017faster}                                                       & \multirow{4}{*}{ResNet-101} & 35.4          & \textbf{56.1} & 37.9          & \textbf{14.7} & 39.0          & 51.4          & 48.08    & 253       \\ \cline{3-5} \cline{7-14} 
                              &                              & 16  & 6e-5 & KL-RPN                                                    &                             & \textbf{35.6} & 55.4          & \textbf{38.3} & 14.3          & \textbf{39.2} & \textbf{52.4} & 48.09    & 260       \\ \cline{1-5} \cline{7-14} 
\multirow{2}{*}{R-FCN}        & \multirow{2}{*}{trainval}    & 8  & 1e-3 & RPN\cite{dai2016r}                                                       &                             & 29.9          & 51.9          & -             & 10.8          & 32.8          & 45.0          & -        & -         \\ \cline{3-5} \cline{7-14} 
                              &                              & 8  & 4e-5 & KL-RPN                                                    &                             & \textbf{31.9} & \textbf{53.9} & \textbf{33.4} & \textbf{13.1} & \textbf{35.0} & \textbf{45.9} & -        & -         \\ \hline
\end{tabular}}\end{center}\vspace{-4mm}
\caption{Comparison of RPN and KL-RPN in COCO test-dev. In \cite{yang2017faster}, we changed the short image size to 600 and train it from scratch. In R-FCN, we use multi-scale training. In the inference, we selected RoIs with top 1000 objectness scores. Speed is measured with GTX 1080Ti. \textbf{bs}: batch size, \textbf{lr}: learning rate.}
\label{tab:coco-test-dev}
\end{table*}
\vspace{-2mm}

\begin{figure}[htb]

\begin{minipage}[b]{1.0\linewidth}
  \centering
  \centerline{\includegraphics[width=1.0\linewidth]{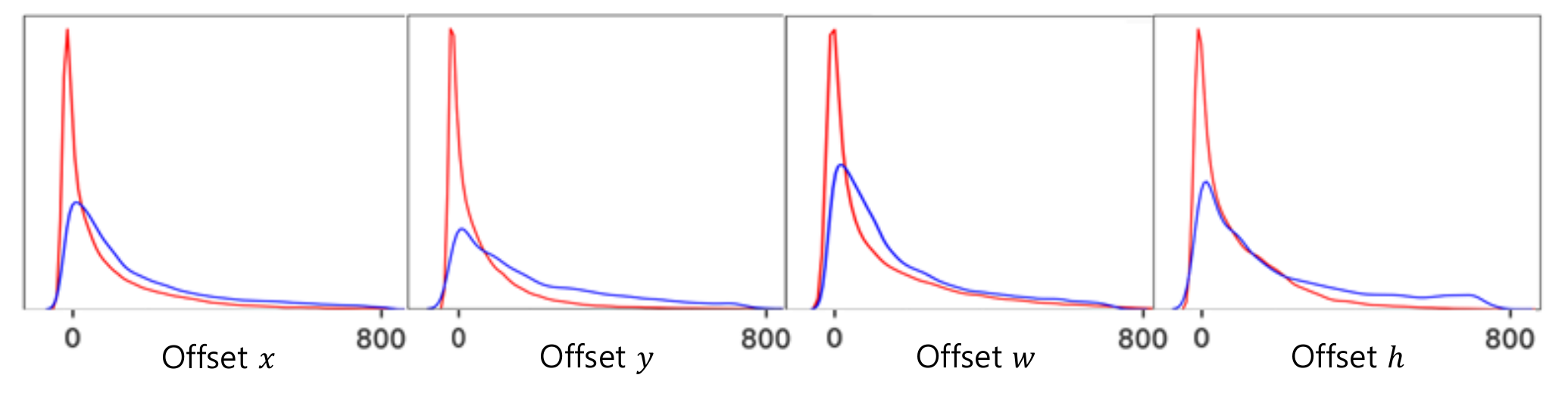}}
  \centerline{(a) RPN}\medskip
\end{minipage}
\begin{minipage}[b]{1.0\linewidth}
  \centering
  \centerline{\includegraphics[width=1.0\linewidth]{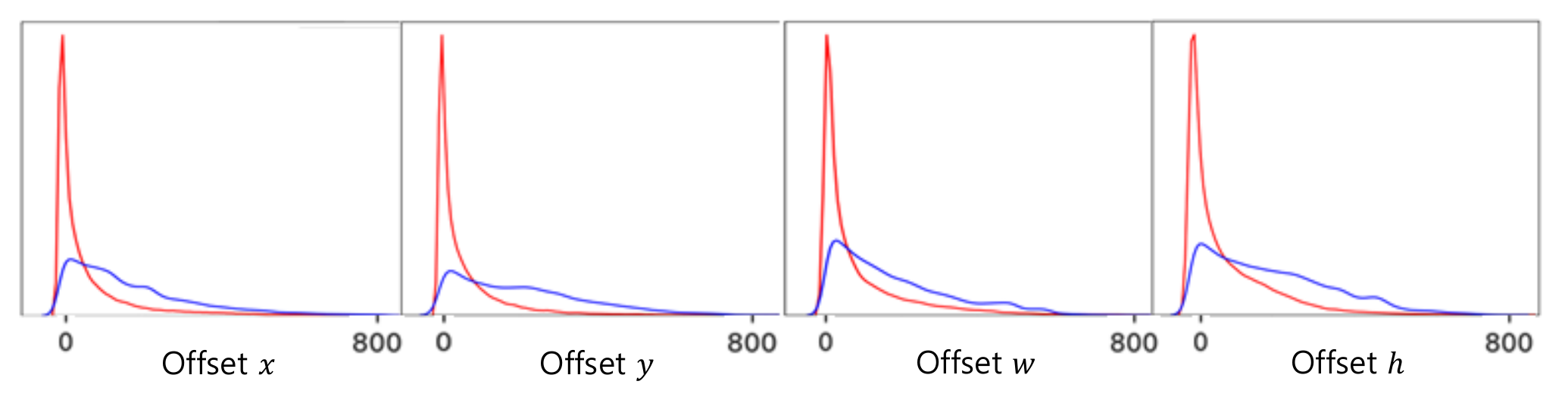}}
  \centerline{(b) KL-RPN}\medskip
\end{minipage}

\vspace{-4mm}
   \caption{The distribution of the $x$, $y$, $w$, and $h$ offsets according to the objectness score. Red, and blue line are high, and low objectness score, respectively. For each image, the number of ground truth bounding boxes $\times$ 5 were sampled in the order of objectness score from high to low, respectively. Base model is VGG-16 with Faster R-CNN. 100 MS COCO minival images were used.}\vspace{-2mm}
\label{fig:ob_score_offset}
\end{figure}
\vspace{-1mm}
\subsection{Training Details}
\label{sec:4_1}
Because of the divergence in the training process, Adam \cite{kingma2014adam} was used as the optimizer with its parameters $\beta_{1}$ and $\beta_{2}$ being 0.9 and 0.999. In addition, the batch sizes are different for each reference experiment. Therefore, different settings were made, and the contents are shown in Table~\ref{tab:coco-test-dev} and Table~\ref{tab:PASCAL}. 
We train the network on 4 Pascal X (Maxwell) GPUs. 
RoI-Align \cite{he2017mask} is adopted in all experiments. The additional parameters of KL-RPN were initialized to $N(0, 0.00001)$. And no gradient clip in VGG-16 backbone. The first convolution layer in 
ResNet-101 are fixed while training. The rest of the experiment setting was conducted in the same manner as \cite{ren2015faster,dai2016r}.

\subsection{Distribution of offset according to the objectness score}
\label{sec:4_2}
In our method, the RoI's objectness score is defined as the uncertainty of the bounding box offset. To show the change in offset from the ground truth bounding box for different objectness scores, we compare the distribution of the offset for high and low objectness scores. 
Among RoIs whose IoU is bigger than zero with ground truth bounding box, Figure~\ref{fig:ob_score_offset} shows the $L_{1}$ offset with ground truth bounding box according to the objectness score for $x$, $y$, $w$, and $h$, respectively. In the Figure~\ref{fig:ob_score_offset} (a), RPN, the offset from the ground truth bounding box has a similar distribution for the bounding box with both high objectness score and low objectness score. It is because the RPN is trained classification and regression independently and does not consider the uncertainty of the bounding box offset while training. Unlike the results of the existing RPN, the distribution of offset change depending on the size of the objectness score in KL-RPN, Figure~\ref{fig:ob_score_offset} (b).

\begin{table}[]
\small
\scalebox{0.87}{
\begin{tabular}{|c|c|c|c|c|c|c|}
\hline
Method                        & \begin{tabular}[c]{@{}c@{}}train\\ data\end{tabular} & bs & lr   & \begin{tabular}[c]{@{}c@{}}Region\\ Proposal\end{tabular} & Backbone                    & mAP           \\ \hline
\multirow{4}{*}{Faster R-CNN} & \multirow{4}{*}{07+12}                               & 1  & 1e-3 & RPN\cite{yang2017faster}                                         & \multirow{2}{*}{VGG-16}     & 75.9          \\ \cline{3-5} \cline{7-7} 
                              &                                                      & 1  & 2e-5 & KL-RPN                                                    &                             & \textbf{77.1} \\ \cline{3-7} 
                              &                                                      & 1  & 1e-3 & RPN\cite{yang2017faster}                                          & \multirow{4}{*}{ResNet-101} & 80.2          \\ \cline{3-5} \cline{7-7} 
                              &                                                      & 1  & 2e-5 & KL-RPN                                                    &                             & \textbf{80.7} \\ \cline{1-5} \cline{7-7} 
\multirow{2}{*}{R-FCN}        & \multirow{2}{*}{07}                                  & 2  & 4e-3 & RPN\tablefootnote{https://github.com/princewang1994/RFCN\underline\,CoupleNet.pytorch}     &                             & 73.8          \\ \cline{3-5} \cline{7-7} 
                              &                                                      & 2  & 4e-5 & KL-RPN                                                    &                             & \textbf{74.6} \\ \hline
\end{tabular}}
\caption{Comparison of RPN and KL-RPN in VOC 2007 test. The short size of input image is fixed to 600. In the inference, we selected RoIs with top 300 objectness scores. \textbf{07}: 07 trainval. \textbf{07+12}: 07 trainval + 12 trainval. \textbf{bs}: batch size, \textbf{lr}: learning rate.}\vspace{-2mm}
\label{tab:PASCAL}
\end{table}

\subsection{Experiments on MS COCO and PASCAL VOC}
\label{sec:4_2}
Table~\ref{tab:coco-test-dev} and Table~\ref{tab:PASCAL} are experiment results. In PASCAL VOC experiment, using KL-RPN increased the mAP, especially in the VGG-16 backbone. In MS COCO experiment, using KL-RPN with Faster R-CNN, there are 2.6 AP improvement in VGG16 and 0.2 AP improvement in ResNet-101. In addition, the detection performance is improved more on the large objects than the medium. Also, the performance is improved without significant change of parameters and inference time. With R-FCN, the results show that the AP increases 2\% compared to the existing RPN, which shows a higher performance improvement than when applied to Faster R-CNN. When we apply our method to the fully convolution method of two-stage object detection, the effect is greater.
\vspace{-1mm}
\section{CONCLUSION}
\label{sec:copyright}

In this paper, we proposed the new region proposal method which defines region proposal problem as one task using KL-Divergence by considering bounding box offset’s uncertainty in the objectness score. The positive sample is considered a Dirac delta function and the negative sample is considered a Gaussian distribution so that the model with Gaussian distribution minimizes the KL-Divergence between them. By using KL-Divergence loss, the network has the advantage of predicting the standard deviation of the offset from the ground truth bounding box and use it as the objectness score. Experiments show that by applying KL-RPN in existing two-stage object detection, the performance is improved and proved that the existing RPN can be replaced successfully.



\bibliographystyle{IEEE}
\bibliography{references}

\end{document}